\documentclass[letterpaper]{article} % DO NOT CHANGE THIS
\usepackage{aaai25}  % DO NOT CHANGE THIS
\usepackage{times}  % DO NOT CHANGE THIS
\usepackage{helvet}  % DO NOT CHANGE THIS
\usepackage{courier}  % DO NOT CHANGE THIS
\usepackage[hyphens]{url}  % DO NOT CHANGE THIS
\usepackage{graphicx} % DO NOT CHANGE THIS
\usepackage{natbib}  % DO NOT CHANGE THIS AND DO NOT ADD ANY OPTIONS TO IT
\usepackage{caption} % DO NOT CHANGE THIS AND DO NOT ADD ANY OPTIONS TO IT
\usepackage{algorithm}
\usepackage{algorithmic}
\usepackage{booktabs} 
\usepackage{amsmath}
\usepackage{amssymb}

\usepackage{newfloat}
\usepackage{listings}
\DeclareCaptionStyle{ruled}{labelfont=normalfont,labelsep=colon,strut=off} % DO NOT CHANGE THIS
\floatstyle{ruled}
\newfloat{listing}{tb}{lst}{}
\floatname{listing}{Listing}
\title{Exploiting Diffusion Prior for Real-World Image Dehazing with Unpaired Training}
\author{
    Yunwei Lan\textsuperscript{\rm 1,2}, Zhigao Cui\textsuperscript{\rm 1 \footnote{Corresponding Author}}, Chang Liu\textsuperscript{\rm 2}, Jialun Peng\textsuperscript{\rm 2}, Nian Wang\textsuperscript{\rm 1}, Xin Luo\textsuperscript{\rm 2}, Dong Liu\textsuperscript{\rm 2 \footnotemark[1]} 
}

\affiliations{
    \textsuperscript{\rm 1}Rocket Force University of Engineering\\
    \textsuperscript{\rm 2}University of Science and Technology of China\\
    yunweilan@mail.ustc.edu.cn, cuizg10@126.com, lc980413@mail.ustc.edu.cn, pjl@mail.ustc.edu.cn, nianwang04@outlook.com, xinluo@mail.ustc.edu.cn, dongeliu@ustc.edu.cn
}

\usepackage{bibentry}
\begin{document}

\maketitle

\begin{abstract}
Unpaired training has been verified as one of the most effective paradigms for real scene dehazing by learning from unpaired real-world hazy and clear images. Although numerous studies have been proposed, current methods demonstrate limited generalization for various real scenes due to limited feature representation and insufficient use of real-world prior.
Inspired by the strong generative capabilities of diffusion models in producing both hazy and clear images, we exploit diffusion prior for real-world image dehazing, and propose an unpaired framework named Diff-Dehazer. Specifically, we leverage diffusion prior as bijective mapping learners within the CycleGAN, a classic unpaired learning framework.
Considering that physical priors contain pivotal statistics information of real-world data, we further excavate real-world knowledge by integrating physical priors into our framework.
Furthermore, we introduce a new perspective for adequately leveraging the representation ability of diffusion models by removing degradation in image and text modalities, so as to improve the dehazing effect.
Extensive experiments on multiple real-world datasets demonstrate the superior performance of our method. Our code https://github.com/ywxjm/Diff-Dehazer. 
\end{abstract}


\section{Introduction}

% Image dehazing is an ill-posed problem in computer vision.
Under hazy conditions, the quality of images is severely degraded.
Such degradation significantly affects the visual appeal of images, and causes information loss, further restricting their performance in other downstream tasks, e.g., object detection \cite{huang2020dsnet}.
Therefore, image dehazing, which aims to restore clear images from hazy ones, has been extensively studied in the past decade.

\begin{figure}[t]
\centering
\includegraphics[width=0.45\textwidth]{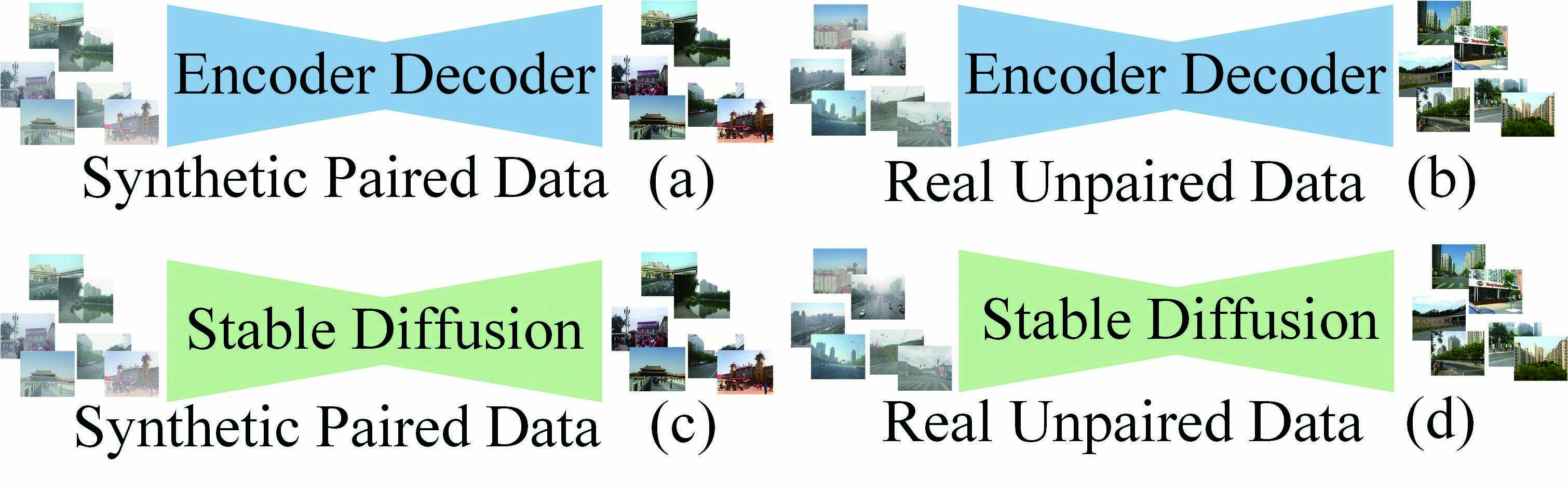} % Reduce the figure size so that it is slightly narrower than the column.
\caption{(a) Previous dehazing methods with paired training. (b) Previous CycleGAN-based dehazing methods with unpaired training. (c) Existing stable diffusion-based dehazing methods with paired training. (d) Our stable diffusion-based dehazing method with unpaired training.}
\label{figure_1}
\end{figure}

Following conventional studies, we normally formulate the hazy image with the Atmospheric Scattering Model (ASM), which can be written as:
\begin{equation} \label{eq: hazy-image-formulation}
    I = Jt + A (1 - t),
\end{equation}
where $I$ represents the hazy image and $J$ refers to its corresponding clear image.
$A$ and $t$ denote the atmospheric light and transmission map, where both of them are usually unknown.
Following the formulation in Eq. (\ref{eq: hazy-image-formulation}), early dehazing methods \cite{he2010single,meng2013efficient} try to estimate $A$ and $t$ with hand-crafted physical priors, and restore hazy images into clear ones by reversing the ASM.
Even so, these methods normally obtain over-saturated results, since hand-crafted physical priors are not universally compatible with all scenes. With the advancement of deep learning, prevailing methods  \cite{zheng2023curricular,song2023vision,qiu2023mb} aim to design neural networks to model physical parameters, or directly restore images in an end-to-end manner, as illustrated in Fig. \ref{figure_1} (a).
These methods primarily utilize synthetic paired data for training since obtaining real-world hazy and clear image pairs is virtually impossible.
Although improved performance is observed, such paired training paradigm often fails to generalize to real-world dehazing scenarios, due to the ill-presenting performance of trained networks that lack real-world information from hazy images.

To tackle the bottleneck of paired training, recent studies \cite{zhao2021refinednet, yang2022self} consider the paradigm of unpaired training for image dehazing, so as to discover available information from real-world hazy images and model the particular mapping between real-world hazy and clear images.
The key problem of unpaired training is how to impose a structure-consistent constraint between hazy and dehazed images, so that the influence of misaligned information caused by unpaired real clear images can be suppressed.
CycleGAN \cite{zhu2017unpaired} offers a classic solution by leveraging a framework to maintain the consistency between the mapped domain and the original one.
Such paradigm, as shown in Fig. \ref{figure_1} (b), is latter adopted by several studies, e.g., CycleDehaze \cite{engin2018cycle}, D4 \cite{yang2022self}, and ODCR \cite{ODCR}, along with promising real-world dehazing performance.
However, these methods struggle to achieve effective representation by a limited number of training images, resulting in sub-optimal performance.
Inspired by the strong representation capabilities of diffusion models, some methods \cite{liu2024diff,lin2024improving} retort to the pre-training and fine-tuning mode, which leverages diffusion prior to improve the dehazing effect, as illustrated in Fig. 1 (c).
Nevertheless, these methods still have limitations in several real-world scenarios due to their reliance on synthetic paired training data that fails to simulate real scenes.

To address the aforementioned problems, we propose an effective paradigm for real-world image dehazing, namely Diff-Dehazer. As shown in Fig. \ref{figure_1} (d), we improve image dehazing with both diffusion prior and the unpaired training paradigm. 
We argue that simply inheriting the unpaired training framework will exhibit sub-optimal dehazing performance without accounting for the physical properties of real-world hazy scenes. Consequently, we conduct a comprehensive investigation on the physical background of image dehazing and integrate physical priors into our framework.   
Furthermore, We observe that text description can improve the resulting image by offering enriched high-level semantics, which has validated its effectiveness in previous studies. Therefore, we adopt a multi-modal paradigm, incorporating text and image to improve the dehazing effect.

Particularly, our contributions are four-fold:
1) We adopt a pre-trained stable diffusion as the foundation of our CycleGAN framework, so as to leverage its strong representation ability in modeling real-world data.
2) We boost the generalization ability of image dehazing with physical priors, whose potentials are greatly neglected in previous studies, and propose Physics-Aware Guidance (PAG) in our framework.
3) We utilize the enriched high-level semantics stored in text modality, and propose Text-Aware Guidance (TAG) to bootstrap textual guidance for image dehazing.
4) To facilitate further studies in this topic, we construct an unpaired real-world dataset, consisting of $6,519$ hazy images and $11,293$ clear images. 
Extensive experiments on existing benchmark datasets illustrate the superior performance of our method compared to state-of-the-art methods.

\section{Related Works}
\subsection{Image Dehazing}
Early dehazing methods extract physical priors from the statistical properties of natural images, and then restore hazy images into clear ones via the ASM.
For example, DCP \cite{he2010single} proposes the dark channel prior to estimate the transmission map and atmospheric light. 
BCCR \cite{meng2013efficient} explores inherent boundary constraint and L1-norm-based contextual regularization to optimize the transmission map.
Nevertheless, these methods are heuristic for real-world image dehazing. They usually lead to over-enhanced results, since hand-crafted priors fail to fit the complexity and diversity of real-world haze.
With the development of deep learning, latter studies concentrate on designing networks for image dehazing.
For instance, C2PNet \cite{zheng2023curricular} customizes a circular learning strategy for image dehazing.
Some methods \cite{guo2022image,song2023vision,qiu2023mb} perform image dehazing based on the Transformer \cite{vaswani2017attention} architecture.
Even so, all the aforementioned methods still struggle to handle several dehazing cases, especially the ones under real-world scenarios, due to their reliance on synthetic paired data. 

% Another mainstream paradigm of image dehazing methods is paired training, which adopts synthetic hazy image and the original clear ones to optimize the dehazing networks, illustrating limited dehazing performance.
To address these limitations, RefineDNet \cite{zhao2021refinednet} designs a weakly supervised two-stage dehazing framework. Based on CycleGAN \cite{zhu2017unpaired}, D4 \cite{yang2022self} introduces a self-augmented dehazing framework to decompose the ASM. Differently, ODCR \cite{ODCR} proposes orthogonal decoupling contrastive regularization to improve the dehazing results.
Although improved performance is observed by these methods, they often fail to demonstrate effectiveness in real-world image dehazing due to insufficient feature representation of conventional encoder-decoder architecture.

\subsection{Diffusion Model-Based Image Restoration}

\begin{figure*}[t]
\centering
\includegraphics[width=1\textwidth]{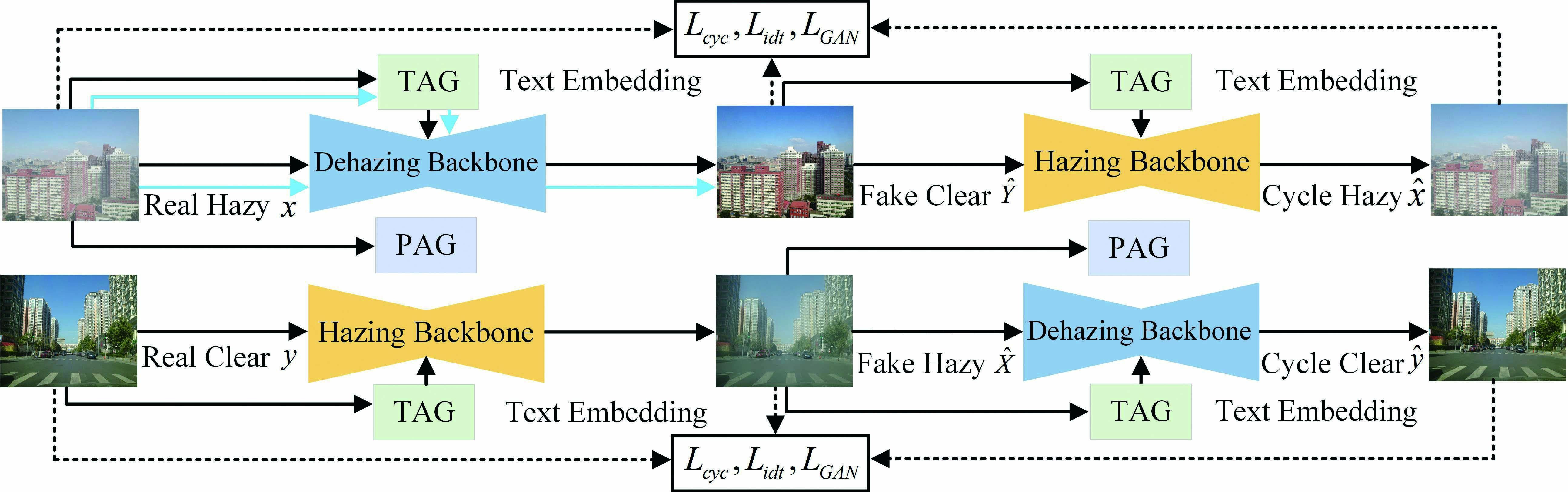} % Reduce the figure size so that it is slightly narrower than the column.
\caption{Overview of our method. Black arrows represent the training process and blue arrows denote the inference process. }
\label{figure_2}
\end{figure*}

Recent advancements in diffusion models \cite{song2021denoising,rombach2022high, liu-2024-etal-stablev2v, liu-etal-2023-lateconstraint, liu-etal-2024-sketchrefiner} have shown superior performance in various vision tasks, including Text-to-Image (T2I) generation \cite{ruiz2023dreambooth} and conditional image generation \cite{zhang2023adding}.
Meanwhile, there is an increasing trend towards the application of diffusion models in low-level vision tasks.
For example, StableSR \cite{wang2023exploring} utilizes prior knowledge encapsulated in the stable diffusion for blind image restoration.
PASD \cite{yang2023pixel} introduces a pixel-aware cross-attention module to enable the stable diffusion to perceive local image structures for image super-resolution.
Relying on the attribute prior in the pre-trained model, PTG-RM \cite{xu2024boosting} designs an additional lightweight module to refine the results of a target restoration network.
For image dehazing, RSHazeDiff \cite{xiong2024rshazediff} proposes a unified Fourier-aware diffusion model for remote sensing image dehazing based on DDPM \cite{ho2020denoising}.
Diff-Plugin \cite{liu2024diff} proposes a lightweight task plugin to provide task-specific priors, guiding the diffusion process for image restoration.  

Despite efforts to leverage off-the-shelf features from diffusion models, the exploration of image dehazing is still insufficient, since all aforementioned methods rely on the use of synthetic data and neglect the vitalness of real-world ones, which fail to deal with real-world hazy images.

\section{Method}
Following the CycleGAN, we establish a hazing-dehazing cycle for unpaired training on real-world data, along with hazing and dehazing processes. 
During the training, the real hazy image $x$ is transformed into a fake clear image and then transformed back into a cycle hazy image $\hat{x}$, as depicted in Fig. \ref{figure_2}. 
A similar process is applied to the real clear image $y$ in reverse order.
Specifically, the hazing process comprises a hazing backbone and Text-Aware Guidance (TAG). 
The dehazing process consists of three components: the dehazing backbone, Physics-Aware Guidance (PAG), and TAG. 
We train the framework using cycle-consistent constraints. After training, we can obtain a clear image from a hazy one using only the dehazing process (see the blue arrows in Fig. \ref{figure_2}).
Details of the aforementioned components are illustrated in the following parts.

\subsection{Backbone Network}
We use SD Turbo (v2.1) \cite{sauer2023adversarial, parmar2024one}, a distilled version of Stable Diffusion (SD) 2.1 as hazing and dehazing backbones, since it allows us to generate a large number of high-quality images within one step.
As shown in Fig. \ref{figure_3} (a), this network comprises three main components: the encoder and decoder of VAE \cite{kingma2013auto}, and the U-Net \cite{ronneberger2015u}.
To efficiently leverage the diffusion prior encapsulated in the SD Turbo, we train it using LoRA \cite{hu2022lora} adapters instead of starting from scratch. Specifically, we only update the input layer of the backbone network as well as the additional LoRA adapters while keeping the remaining model parameters frozen.

Previous SD-based image restoration methods typically begin with mapping the image from pixel to latent space via pre-trained VAEs \cite{vqgan}, where the restoration process is then performed in the VAE latent space.
Once the VAE features are generated by the diffusion model, the restored image is converted back to pixel space via the decoder of VAE.
However, these methods are not directly applicable to image dehazing since performing dehazing in a highly compressed space inevitably leads to significant loss of image information.
Consequently, the final restored images exhibit noticeably lower fidelity and significant discrepancies from the original ones in multiple aspects, e.g., regional details and textures.
To preserve the details of the source image during the dehazing process, we implement a skipped connection between the encoder and decoder of the VAE.

\subsection{Text-Aware Guidance}

Textual feature is proven to offer enriched high-level semantics for generative models, and has already demonstrated its effectiveness in stable diffusion \cite{rombach2022high}.
Based on this finding, we expect to further integrate the textual information into our framework, so as to enhance the dehazing process with textual features.
In doing so, we propose Text-Aware Guidance (TAG), with its illustration shown in Fig. \ref{figure_3} (b).
Specifically, we first employ a pre-trained image captioner, i.e., BLIP-2 \cite{li2023blip}, to produce a caption for the input image, where the extracted caption is then utilized in further processes.
Note that users can manually customize the text description according to their needs when inferencing any real-world hazy images. 

Different from the explicit use of text descriptions in current methods, we introduce TAG from the following two perspectives.
On the one hand, we refine these captions by eliminating haze-related terms, e.g., ``\textit{haze}'', ``\textit{fog}'', etc., so as to explicitly facilitate image dehazing in the text modality. 
On the other hand, we leverage both positive and negative prompts via classifier-free guidance \cite{ho2021classifierfree} to improve the quality of the dehazed image, with the refined caption serving as the positive prompt.
Considering that the hazing attribute might be too complicated to describe with explicit words, we learn a prompt by textual inversion \cite{gal2022image} from thousands of real-world hazy images to holistically depict the hazing attribute implicitly. 
With acquired positive and negative prompts, we achieve more comprehensive semantic guidance via the text encoder of CILP and further boost the image quality. 
As for the hazing process, we first obtain the image caption directly through BLIP-2, and then combine the haze-related terms with the extracted caption, considering it as a positive prompt.
For the negative prompt, we assign an empty text.

\subsection{Physics-Aware Guidance}

\begin{figure*}[t]
\centering
\includegraphics[width=1\textwidth]{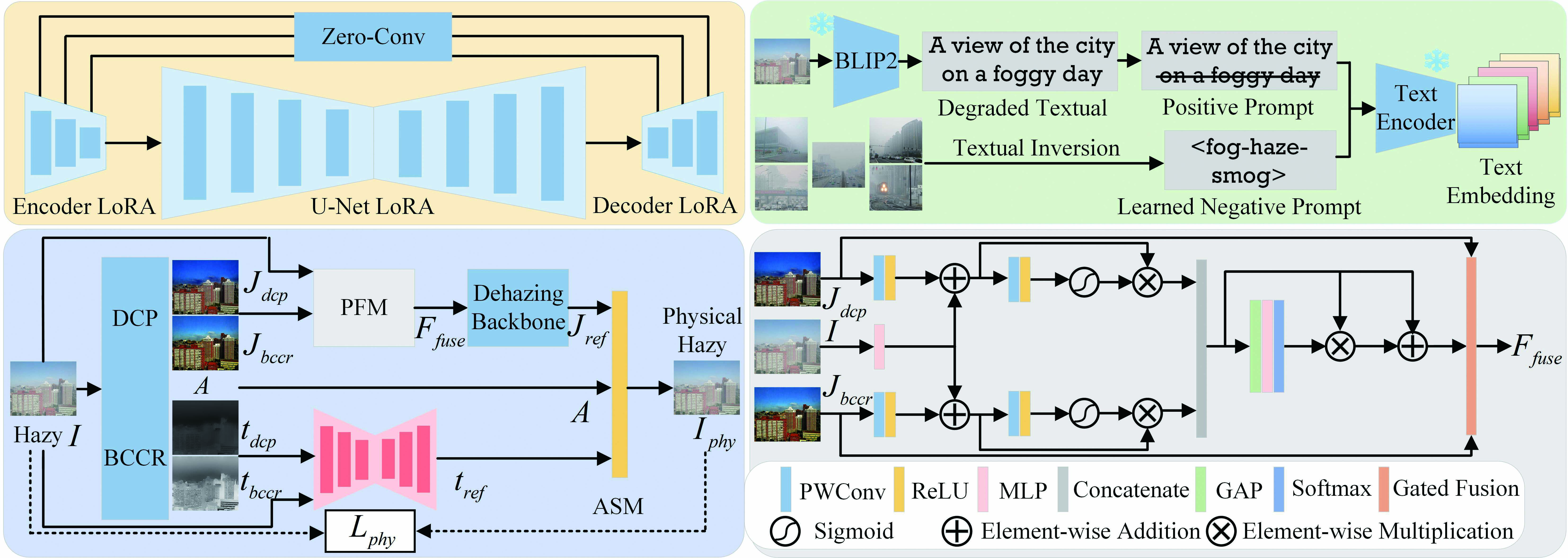} % Reduce the figure size so that it is slightly narrower than the column.
\caption{Orange Area: Backbone network of our framework. Note that hazing and dehazing backbones share the same U-Net and employ two individual VAEs. Green Area: Text-Aware Guidance (TAG). Blue Area: Physics-Aware Guidance (PAG). Dark Channel Prior (DCP) and Boundary Constraint and Contextual Regularization (BCCR) are two physical prior-based dehazing methods. ASM is the Atmospheric Scattering Model. Gray Area: The structure of Perceptual Fusion Model (PFM).}
\label{figure_3}
\end{figure*}
Previous studies have demonstrated that using physical priors is more likely to remove haze, particularly in real-world scenarios since these priors are statistical laws from a large number of real-world clear images.
Therefore, we integrate physical priors into our framework and introduce Physics-Aware Guidance (PAG) for image dehazing, as illustrated in Fig. \ref{figure_3} (c).
Different from previous methods, we investigate various physical priors and integrate two well-performing (i.e., DCP and BCCR) into the framework.
After obtaining preliminary dehazed result ${J}$, atmospheric light ${A}$, and transmission map ${t}$ in Eq. (\ref{eq: hazy-image-formulation}), we reconstruct the hazy image via the ASM and design a physical loss.
By fine-tuning the dehazing backbone, the model adheres to the underlying physical principles, achieving more effective dehazing and physical awareness. We provide more details on how we obtain preliminary results in our supplementary materials.

\subsubsection{Reconstruction of Hazy Image}

To leverage the physical priors encapsulated in clear images dehazed by DCP and BCCR, we treat them as clear images for hazy image reconstruction, thereby compelling the model to learn more about the physical properties of real-world haze.
Observing that either ${J}_{dcp}$ or ${J}_{bccr}$ is better than the other in some regions under various application scenarios, we further refine them rather than applying them directly.
Specifically, for ${t}_{dcp}$ and ${t}_{bccr}$, we concatenate and feed them into a U-Net to obtain a refined transmission map ${t}_{ref}$.
Given that the input image contains source information, we take it as input and feed it into the U-Net along with the transmission map.
To combine the advantages of ${J}_{dcp}$ and ${J}_{bccr}$, we introduce the Perceptual Fusion Model (PFM) to effectively fuse them, resulting in a composite image ${F}_{fuse}$.
Within our framework, the dehazing backbone is capable of not only restoring clear images from hazy inputs but also enhancing the image quality of clear inputs.
Consequently, we consider the dehazing backbone as a refined network and feed ${J}_{fuse}$ into it to obtain a refined clear image ${J}_{ref}$ with enhanced natural details and physical awareness.
In doing so, we can reconstruct a more qualified hazy image ${I}_{phy}$ using refined ${J}_{ref}$, ${t}_{ref}$, and $A$, imposing a more reasonable physical constraint.

The structure of the PFM is shown in Fig. \ref{figure_3} (d).
We utilize point-wise convolution layers, ReLU, and MLP to extract latent features from ${J}_{dcp}$, ${J}_{bccr}$, and ${I}$.
To preserve any information potentially lost in the initial dehazed images, we add the feature of ${I}$ to those of ${J}_{dcp}$ and ${J}_{bccr}$.
Guided features are obtained using point-wise convolution layers, ReLU, Sigmoid, and residual connections.
Subsequently, we concatenate and re-weight them using global average pooling, MLP, and Softmax. The concatenated features are then multiplied by the obtained weight, followed by a residual connection to produce coarse-fused features. Finally, we re-weight the proportions of ${J}_{dcp}$ and ${J}_{bccr}$ via a gated fusion \cite{chen2019gated}, resulting in fine-fused results ${J}_{fuse}$.

\section{Loss Function}

\begin{figure*}[t]
\centering
\includegraphics[width=1\textwidth]{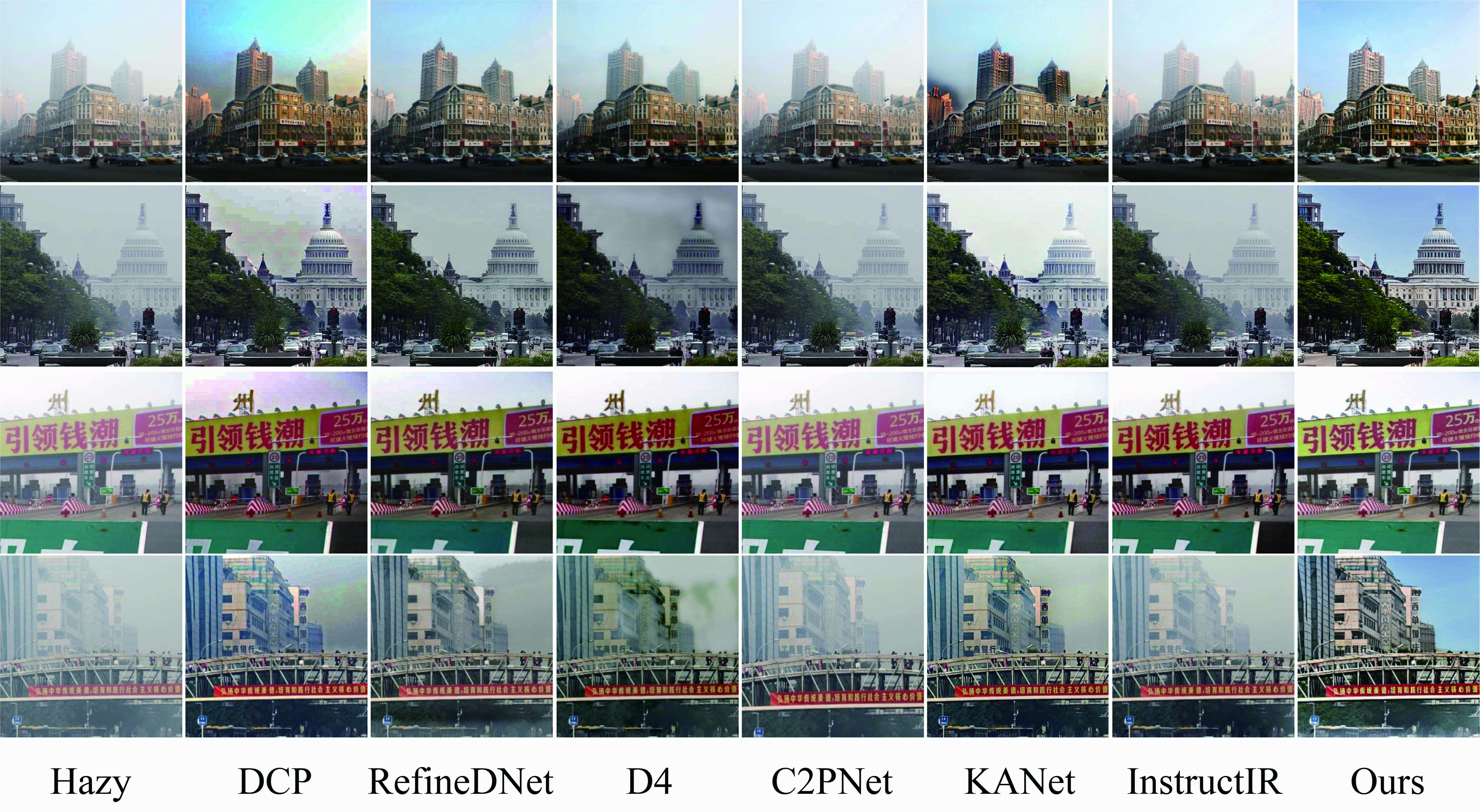} % Reduce the figure size so that it is slightly narrower than the column.
\caption{Visual comparison of samples from Haze2020 and RTTS. Our method can effectively remove haze and generate high-quality images with natural color and realistic contrast. More visual results are presented in our supplementary materials.}
\label{figure_4}
\end{figure*}

\begin{table*}[t]
\centering
\small
\begin{tabular}{ccccccccc}
\toprule
 & \multicolumn{4}{c}{RTTS} & \multicolumn{4}{c}{Haze2020} \\ \midrule
 & FID $\downarrow$  & NIQE $\downarrow$ & MUSIQ $\uparrow$ & CLIPIQA $\uparrow$ & FID $\downarrow$ & NIQE $\downarrow$ & MUSIQ $\uparrow$ & CLIPIQA $\uparrow$ \\ \midrule
DCP & 73.017 & 4.271 & 52.935 & 0.276 & 91.948 & 3.827 & 54.375 & 0.364 \\
BCCR & 75.984 & 4.289 & 52.429 & 0.270 & 92.798 & 3.765 & 54.438 & 0.362 \\
RefineDNet (TIP2021) & 65.076 & 4.012 & 56.117 & 0.257 & 88.229 & 3.972 & 54.779 & 0.350 \\
PSD (CVPR2021) & 73.847 & 4.017 & 56.430 & 0.272 & 91.922 & 3.868 & 58.522 & 0.257 \\
Dehamer (CVPR2022) & 66.208 & 4.882 & 53.787 & 0.365 & 84.416 & 4.118 & 54.528 & 0.419 \\
D4 (CVPR2022) & 69.400 & 4.637 & 58.445 & 0.351 & 87.536 & 3.971 & 53.458 & 0.309 \\
DehazeFormer (TIP2023) & 68.636 & 4.693 & 53.692 & 0.352 & 82.842 & 4.208 & 55.590 & 0.418 \\
C2PNet( CVPR2023) & 66.117 & 5.037 & 53.960 & \textbf{0.390} & 83.959 & 4.206 & 54.565 & 0.423 \\
KANet( TPAMI2024) & 64.963 & 4.339 & 54.513 & 0.285 & 84.888 & 3.742 & 56.411 & 0.376 \\
InstructIR (ECCV2024) & 66.278 & 4.902 & 54.464 & 0.369 & 83.561 & 4.164 & 55.080 & 0.422 \\
Diff-Plugin (CVPR2024) & 65.787 & 5.334 & 50.740 & 0.361 & 80.965 & 4.407 & 52.752 & 0.424 \\
Ours & \textbf{52.344} & \textbf{3.943} & \textbf{59.207} & 0.328 & \textbf{71.984} & \textbf{3.571} & \textbf{62.273} & \textbf{0.439} \\ \bottomrule
\end{tabular}
\caption{Quantitative results on RTTS and Haze2020. The best results are denoted in \textbf{bold}.}
\label{table_2}
\end{table*}

\begin{figure*}[t]
\centering
\includegraphics[width=1\textwidth]{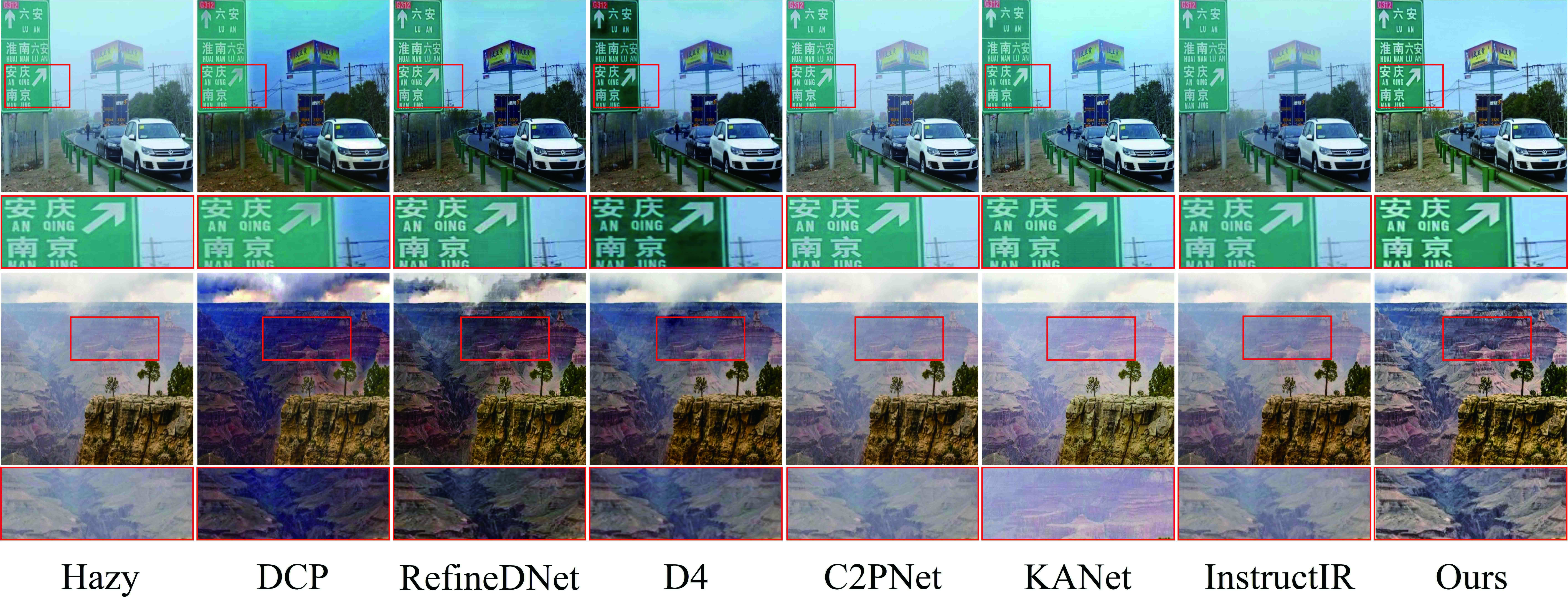} % Reduce the figure size so that it is slightly narrower than the column.
\caption{Visual comparison of samples from Haze2020. Areas where our method works better are boxed out and zoomed in. Our method can generate clear images with high fidelity and discriminative textures.}
\label{figure_5}
\end{figure*}

\begin{table}[t]
\centering
\small
\begin{tabular}{ccccc}
\toprule
 & \multicolumn{4}{c}{OHAZE} \\ \midrule
 & PSNR $\uparrow$ & SSIM $\uparrow$ & LPIPS $\downarrow$ & VSI $\uparrow$ \\ \midrule
DCP & 17.005 & 0.819 & 0.242 & 0.945  \\
BCCR & 15.487 & 0.753 & 0.261 & 0.943  \\
RefineDNet & 18.693 & 0.755 & 0.260 & \textbf{0.954} \\
PSD & 14.727 & 0.717 & 0.316 & 0.930 \\
Dehamer & 17.827 & 0.688 & 0.330 & 0.927  \\
D4 & 16.767 & 0.690 & 0.290 & 0.939 \\
DehazeFormer & 15.243 & 0.672 & 0.339 & 0.911 \\
C2PNet & 18.014 & 0.708 & 0.317 & 0.929 \\
KANet & 17.713 & 0.814 & 0.265 & 0.939 \\
InstructIR & 18.616 & 0.716 & 0.321 & 0.931 \\
Diff-Plugin & 16.470 & 0.526 & 0.364 & 0.920 \\
Ours & \textbf{19.539} & \textbf{0.825} & \textbf{0.195} & \textbf{0.954} \\ \bottomrule
\end{tabular}
\caption{Quantitative results on OHAZE. The best results are denoted in \textbf{bold}.}
\label{table_3}
\end{table}

\begin{figure*}[t]
\centering
\includegraphics[width=1\textwidth]{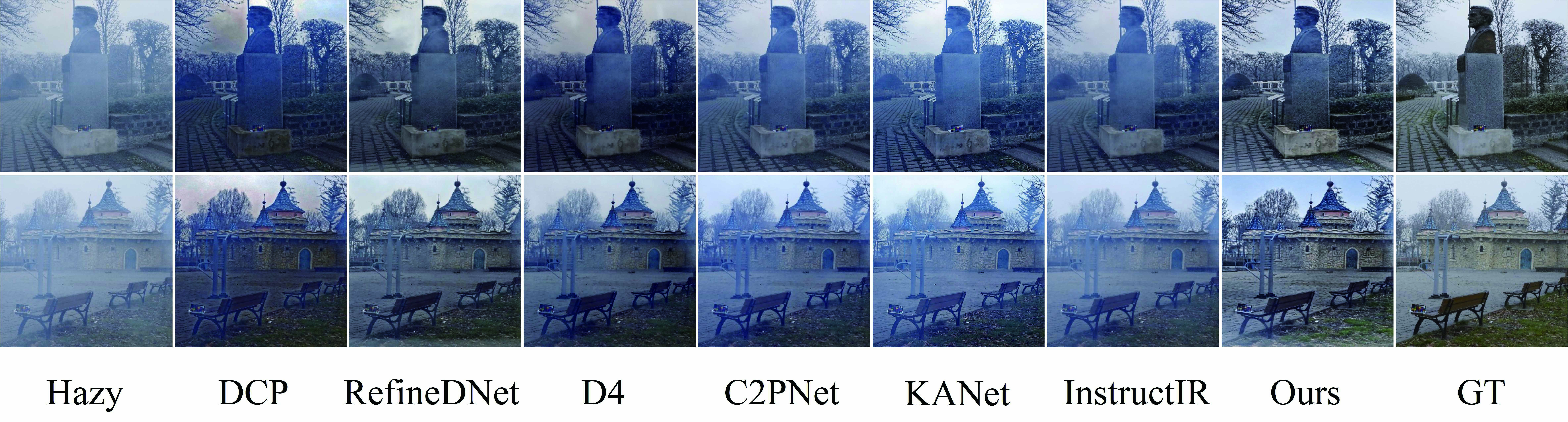} % Reduce the figure size so that it is slightly narrower than the column.
\caption{Visual comparison of samples from OHAZE.}
\label{figure_6}
\end{figure*}
% \begin{equation}
% \arg \min_{G}{L_{cyc}+\lambda_{phy}L_{phy} +\lambda _{idt}L_{idt}+\lambda _{GAN}L_{GAN}},
% \label{eq_16}
% \end{equation}
Following CycleGAN \cite{zhu2017unpaired}, we design the training loss for the proposed model as:
\begin{equation}
L = {L_{cyc}+\lambda_{phy}L_{phy} +\lambda _{idt}L_{idt}+\lambda _{GAN}L_{GAN}},
\label{eq_16}
\end{equation}
where $L_{cyc}$, $L_{phy}$, $L_{idt}$, and $L_{GAN}$ represent the cycle consistency loss, prior loss, identity loss, and GAN loss, respectively.
$\lambda_{phy}$, $\lambda_{idt}$, and $\lambda_{GAN}$ are the corresponding hyper-parameters to control the weight of $L_{phy}$, $L_{idt}$, and $L_{GAN}$. We present more details of $L_{cyc}$, $L_{idt}$, and $L_{GAN}$ in our supplementary materials since they are similar to CycleGAN. 

\subsubsection{Physical Loss}
For our model, due to the use of physical priors, we design the physical loss as:
\begin{equation}
{L}_{phy} = {L}_{rec}(I, {I}_{phy}),
\label{equation_10}
\end{equation}
where ${I}_{phy}={J}_{ref}{t}_{ref}+A(1-{t}_{ref})$ refers to the reconstructed hazy image derived by the ASM. $I$ denotes the overall notation involving both real/fake hazy images.
${L}_{rec}$ is the combined distance of ${L}_{1}$ and LPIPS \cite{zhang2018unreasonable}.

\section{Experiments and Discussions}

\subsection{Datasets}
We collect over $7,000$ real-world hazy images from RESIDE \cite{li2018benchmarking}, and select over $10,000$ clear images from ADE20K \cite{zhou2019semantic} as well as OTS (a subset of RESIDE) to construct hazy and clear images as our training set, respectively.

To evaluate the effectiveness of our proposed method, we conduct qualitative and quantitative experiments on various real-world image datasets, including URHI, RTTS, Haze2020, OHAZE \cite{ancuti2018haze}, NHAZE \cite{ancuti2020nh}, and the proposed dataset in \citet{fattal2014dehazing}. Specifically, URHI and RTTS are two subsets of RESIDE that contain over $4,000$ real-world hazy images. Haze2020 consists of over $1,000$ hazy images selected by DA \cite{shao2020domain}. OHAZE and NHAZE contain 45 and 55 pairs of outdoor hazy images, respectively, which are artificially generated using a haze machine.
The dataset proposed in \citet{fattal2014dehazing} contains 37 real-world hazy images. 

% Notably, we choose outdoor images for training and testing, since the pre-trained SD Turbo and the CLIP inherently assume haze occurrences to be exclusive to outdoor environments.
% Using indoor images would contradict this assumption and potentially degrade dehazed results.

\subsection{Comparisons with State-of-the-Art Methods}

We compare the performance of our method with several state-of-the-art methods, including DCP \cite{he2010single}, BCCR \cite{meng2013efficient}, RefineDNet \cite{zhao2021refinednet}, PSD \cite{chen2021psd}, Dehamer \cite{guo2022image}, D4 \cite{yang2022self}, DehazeFormer \cite{song2023vision}, C2PNet \cite{zheng2023curricular}, KANet \cite{feng2024kanet}, InstructIR \cite{conde2024high}, and Diff-Plugin \cite{liu2024diff}. Notably, DCP and BCCR are physical prior-based methods. RefineDNet and D4, along with our method, do not require paired data for training and are categorized as weakly supervised methods. The remaining methods are fully supervised methods and necessitate paired data for training. For images with available ground truth, we evaluate these methods using full-reference metrics such as PSNR, SSIM \cite{wang2004image}, LPIPS, and VSI \cite{zhang2014vsi}. In cases where ground truth is unavailable, we evaluate them by no-reference metrics, including FID \cite{heusel2017gans}, CLIPIQA \cite{wang2023exploring}, NIQE \cite{mittal2012making}, and MUSIQ \cite{ke2021musiq}.

\subsubsection{Results on RTTS and Haze2020.}
 Fig. \ref{figure_4} shows a qualitative comparison of the results on RTTS and Haze2020. The physical prior-based method DCP effectively processes the images. However, due to the inherent limitations of physical priors, it results in the inevitable over-enhancement of the images, producing darker results with lower visual quality. Unpaired dehazing methods, such as RefineDNet and D4, can dehaze to a certain extent as they do not require paired synthetic datasets for training. However, these methods fail to thoroughly dehaze due to limited feature representation and insufficient use of physical priors, resulting in images with residual artifacts and insufficient details. MB-TaylorFormer and C2PNet demonstrate exceptional performance on synthetic images through well-designed networks and training strategies, but they still face challenges with real-world hazy images and even fall short of the efficacy of a physical prior-based method. Similarly, recently proposed all-in-one image restoration methods have only been demonstrated to be effective on synthetic images. When dealing with real-world hazy images, these methods (e.g., Instruct IR) are almost ineffective. The proposed method outperforms other state-of-the-art methods and can generate more realistic and natural images with rich details and textures, as well as high contrast.

The quantitative results of RTTS and Haze2020 are shown in Table \ref{table_2}. In RTTS, our method exhibits superior performance in FID, NIQE, and MUSIQ scores. For CLIPIQA, our method is ranked lower. This can be attributed to the integration of physical priors. However, our intention of using physical priors is to limit the stochastics of the stable diffusion to some extent, ensuring that the resulting dehazed image is more natural while not generating something that does not match the original hazy image. Therefore, it is justifiable to sacrifice a certain amount of CLIPIQA. In Haze2020, our method surpasses others across various metrics.

To further illustrate the superior performance of our proposed method, we zoom in on local details within selected images and show them in Fig. \ref{figure_5}. As we can see, our method not only effectively restores the textures of images but also maintains the image fidelity.

\subsubsection{Results on OHAZE}
In Table \ref{table_3} and Fig. \ref{figure_6}, we present the qualitative and quantitative results of OHAZE. Note that we do not need to retrain the network when evaluating this dataset. 
Our method exhibits superior performance in visualization and quantitative metrics compared to other methods, further highlighting its strong generalization ability, as it can effectively process real-world hazy images with various types and haze densities, producing high-quality results. 

The results of NHAZE, URHI, and the dataset proposed in \citet{fattal2014dehazing} are presented in our supplementary materials. Moreover, We compare model efficiency and object detection accuracy in our supplementary materials.   

\subsection{Ablation Study}
\begin{figure}[t]
\centering
\includegraphics[width=0.45\textwidth]{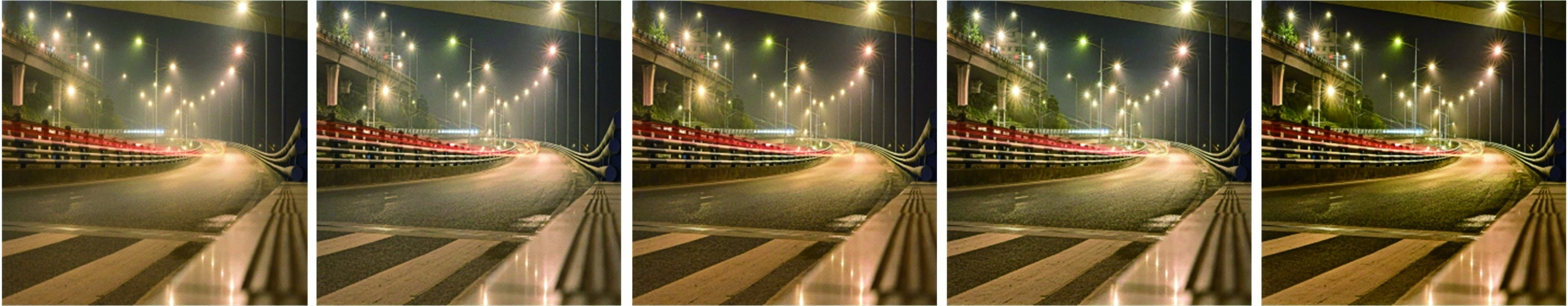} % Reduce the figure size so that it is slightly narrower than the column.
\caption{Ablation study of each component. From left to right, respectively, the hazy image, the result of \textbf{Backbone}, the result of \textbf{Backbone+PAG}, the result of \textbf{Backbone+TAG}, and the result of \textbf{Ours}.}
\label{figure_7}
\end{figure}

We conduct a series of ablation experiments to verify the effectiveness of each key component. Specifically, we discuss the backbone network, PAG, and TAG.
We first establish four variants: \textbf{Backbone:} Removing TAG and PAG. \textbf{Backbone + PAG:} Removing TAG. \textbf{Backbone + TAG:} Removing PAG. \textbf{Ours:} The full model of our method. As shown in Table \ref{table_4} and Fig. \ref{figure_7}, the proposed method achieves the best performance in metrics and visual appeals, which validates that each component plays a critical role in our framework.
Moreover, we conduct an ablation study to analyze the effectiveness of the backbone network supplementary materials.

\begin{table}[t]
\centering
\small
\begin{tabular}{ccccc}
\toprule
Backbone & \checkmark & \checkmark & \checkmark & \checkmark \\
PAG & & \checkmark & & \checkmark \\
TAG & & & \checkmark & \checkmark \\ \midrule
FID & 74.575 & 73.096 & 74.100 & \textbf{71.984} \\
NIQE & 3.783 & 3.762 & 3.574 & \textbf{3.571} \\
MUSIQ & 60.606 & 60.571 & 61.882 & \textbf{62.273} \\
CLIPIQA & 0.427 & 0.420 & 0.436 & \textbf{0.439} \\ \bottomrule
\end{tabular}
\caption{Ablation study of each component.}
\label{table_4}
\end{table}

\begin{figure}[t]
\centering
\includegraphics[width=0.36\textwidth]{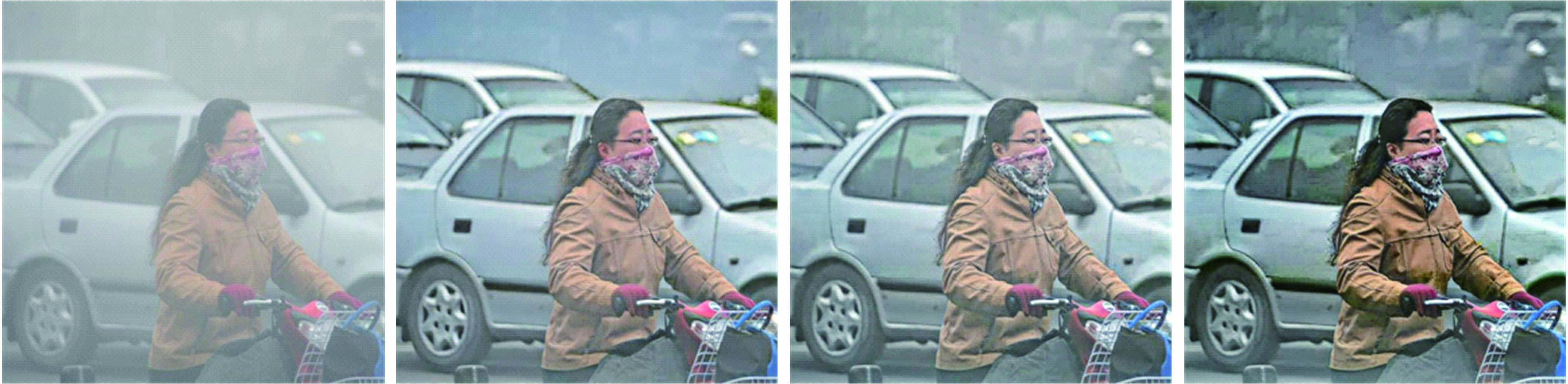} % Reduce the figure size so that it is slightly narrower than the column.
\caption{Ablation study of positive and negative prompts. From left to right, respectively, the hazy image, the result with no prompt, the result with only a positive prompt, and the result with both positive and negative prompts.}
\label{figure_8}
\end{figure}

\begin{table}[t]
\centering
\small
\begin{tabular}{cllcc}
\toprule
Baseline & \multicolumn{1}{c}{\checkmark} & \multicolumn{1}{c}{\checkmark} & \checkmark & \checkmark \\
DCP &  & \multicolumn{1}{c}{\checkmark} &  & \checkmark \\
BCCR &  &  & \checkmark & \checkmark \\ \midrule
FID & 74.100 & 76.768 & \multicolumn{1}{l}{74.999} & \textbf{71.984} \\
NIQE & 3.574 & \textbf{3.534} & 3.574 & 3.571 \\
MUSIQ & 61.882 & 61.585 & \multicolumn{1}{l}{61.399} & \textbf{62.273} \\
CLIPIQA & 0.436 & 0.414 & \multicolumn{1}{l}{0.420} & \textbf{0.439} \\ \bottomrule
\end{tabular}
\caption{Ablation study of physical priors.}
\label{table_5}
\end{table}

% \begin{table}[t]
% \centering
% \begin{tabular}{ccccc}
% \hline
% Variants & A & B & C & D \\ \hline
% Backbone & \checkmark & \checkmark & \checkmark & \checkmark \\
% PAG & & \checkmark & & \checkmark \\
% TAG & & & \checkmark & \checkmark \\ \hline
% FID & 74.575 & 73.096 & 74.100 & \textbf{71.984} \\
% NIQE & 3.783 & 3.762 & 3.574 & \textbf{3.571} \\
% MUSIQ & 60.606 & 60.571 & 61.882 & \textbf{62.273} \\
% CLIPIQA & 0.427 & 0.420 & 0.436 & \textbf{0.439} \\ \hline
% \end{tabular}
% \caption{Ablation study of each component}
% \label{table_4}
% \end{table}

% 首先画一个表进行总的分析然后分析各个模块就在hazy2020上测试,四个指标
% 画一个对照图

\subsubsection{Analysis of Text-Aware Guidance}
We conduct an ablation study to validate how TAG works within our method. Firstly, we evaluate the effectiveness of positive and negative prompts. As illustrated in Fig. \ref{figure_8}, using a positive text prompt alleviates the generation of mismatched details during the dehazing process, which is caused by the inherent stochastics of the stable diffusion. The employment of a negative text prompt can further improve the dehazing effect.
% 加一组实验，用paired方式进行训练
% 给指标
Additionally, we demonstrate the impact of the guidance scale and present the results in our supplementary materials. 
% 折线图

\subsubsection{Analysis of Physics-Aware Guidance}
We demonstrate how physical priors in our framework affect the results. We establish some variants as follows: \textbf{Baseline: } the full model of our method without PAG. \textbf{Baseline + DCP:} Removing the use of BCCR. \textbf{Baseline + BCCR:} Removing the use of DCP. \textbf{Ours:} We adopt both DCP and BCCR to constrain the network training. Table \ref{table_5} demonstrates that our method yields superior results from a comprehensive perspective. Additionally, we validate the impact of the weight of physical loss and provide the results in our supplementary materials.

\section{Conclusion}
In this paper, we delve into the potential of prior encapsulated in the stable diffusion and physical priors derived from natural images, thereby proposing an unpaired framework for real-world image dehazing named Diff-Dehazer. Furthermore, by leveraging enriched high-level semantics contained in text, we perform dehazing in text and image modalities to get more qualified results. Extensive experiments have validated the superiority of our method. However, due to our fine-tuning paradigm, our method may suffer from the inherent problems of diffusion models, which interact with noises and may produce diversified images. This may cause misalignment in some cases with severe haze. Despite this, we perform stably on most cases.

\bibliography{aaai25}

\end{document}